\documentclass[journal]{IEEEtran}

\usepackage{amsmath}
\usepackage{enumerate}
\usepackage{mathrsfs}
\usepackage{makeidx}         
\usepackage{graphicx}        
\usepackage{multicol}        
\usepackage{subfigure}
\usepackage{epsfig,amssymb,latexsym}
\usepackage{psfrag}

\usepackage{epsfig,amssymb,latexsym}
\usepackage{psfrag}
\usepackage[ruled,vlined]{algorithm2e}

\usepackage{algorithm2e}

\usepackage{color}

\def\QED{~\rule[-1pt]{5pt}{5pt}\par\medskip}

\newcommand{\bd}{\begin{displaymath}}
\newcommand{\ed}{\end{displaymath}}
\newcommand{\be}{\begin{equation}}
\newcommand{\ee}{\end{equation}}
\newcommand{\bea}{\begin{eqnarray}}
\newcommand{\eea}{\end{eqnarray}}
\newcommand{\ba}{\begin{array}}
\newcommand{\ea}{\end{array}}
\newcommand{\bc}{\begin{center}}
\newcommand{\ec}{\end{center}}

\newcommand{\nb}{\nonumber}

\newcommand{\re}[1]{(\ref{#1})}

\begin{document}
%
\title{Improving Tracking through Human-Robot Sensory Augmentation}

\author{Yanan Li*,~\IEEEmembership{Member,~IEEE}, Jonathan Eden*, Gerolamo Carboni*,
        and~Etienne Burdet,~\IEEEmembership{Member,~IEEE}
\thanks{This research was supported by the European Commission grant EU-H2020 COGIMON (644727) and UK EPSRC grants EP/NO29003/1, EP/T006951/1.}
\thanks{*: The first three authors contributed equally to the work.}
\thanks{Y. Li, J. Eden, G. Carboni and E. Burdet are or were with the Department of Bioengineering, Imperial College of Science, Technology and Medicine, SW7 2AZ London, UK.} 
\thanks{Y. Li is with the Department of Engineering and Informatics, University of Sussex, Brighton, BN1 9RH, UK. Email: yl557@sussex.ac.uk}
}

\maketitle

\begin{abstract}
This paper introduces \emph{human-robot sensory augmentation} and illustrates it on a tracking task, where performance can be improved by the exchange of sensory information between the robot and its human user. It was recently found that during interaction between humans, the partners use each other's sensory information to improve their own sensing, thus also their performance and learning \cite{Takagi2017}. In this paper, we develop a computational model of this unique human ability, and use it to build a novel control framework for human-robot interaction. The human partner's control is formulated as a feedback control with unknown control gains and desired trajectory. A Kalman filter is used to estimate first the control gains and then the desired trajectory. The estimated human partner's desired trajectory is used as augmented sensory information about the system and combined with the robot's measurement to estimate an uncertain target trajectory. Simulations and an implementation of the presented framework on a robotic interface validate the proposed observer-predictor pair for a tracking task. The results obtained using this robot demonstrate how the human user's control can be identified, and exhibit similar benefits of this sensory augmentation as was observed between interacting humans.
\end{abstract}

\IEEEpeerreviewmaketitle

\section{Introduction}
Teleoperation (as used e.g. in the surgical robot) implements control according to a strict hierarchy, where the slave robot follows the movement or force imposed by a human master \cite{Gupta2006,Xia2011,Taylor2016}. At the other extreme, rehabilitation robots used in physical therapy generally impose a movement to a human user's limbs \cite{Yu2015,Colombo2018}. However, more egalitarian task sharing between a robot and its human user may be developed to take the opportunities offered by their interaction \cite{Mortl2012,Jarrasse2014,Losey2018}. In fact, there is an increasing interest in shared control, where the robot and human control the same system simultaneously \cite{Chipalkatty2013,Palunko2014, Na2017}. In the framework of \cite{Jarrasse2012, Li2019}, a human and a robot can interact according to roles defined by specific cost functions using e.g. game theory to compute the motor commands. In this manner, the robot can carry out a predefined regular task while the human intervenes when needed \cite{Uri2006, Li2015TRO}.

How to collaborate with a human? It is often stated that collaborative strategies should be designed so that the robot and human use the best of their respective capabilities. By this it is usually meant that the robot would carry heavy loads according to targets identified by the human user, who has the superior analysis and sensorimotor intelligence capabilities \cite{Rozo2016,Salmi2018}. A lot of research on impedance control falls into this category, e.g. \cite{Al-Jarrah1997, Albu-SchafferA07IJRR,Erden2015}. While the aforementioned works focus on how humans and robots can share the task load and control effort, we propose here a different strategy according to which \emph{a human and its robot could exchange haptic information during physical interaction to complement their own sensing}.

\begin{figure}[thpb]
\centering
\includegraphics[width=0.9\columnwidth]{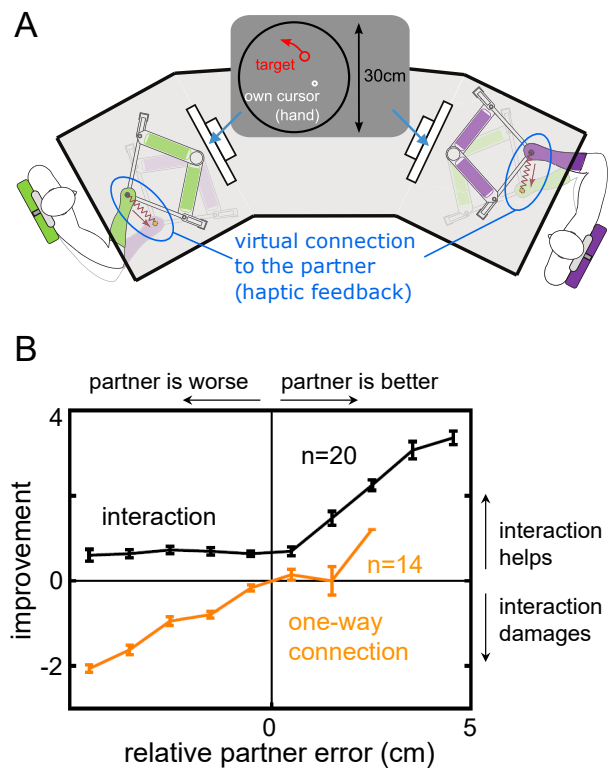}
\caption{Human-human experiment from \cite{Ganesh2014}: When two humans track a common randomly moving target while being connected through a (virtual) elastic band (A), they improve the tracking so that even the best partner (according to individual performance) benefits from the interaction with the worse one (B). It has been shown in \cite{Takagi2017} that these benefits are due to \emph{haptic communication} between the partners, where the partners understand each other's motion goal and integrate this information to improve own visual tracking.}
\label{f:Ganesh2014}
\end{figure}
This \emph{sensory augmentation} is in line with the notion of the observation-control duality in control theory \cite{Kalman1960} but has not been studied for human-robot collaboration. Interestingly, it has been observed that when humans are in physical contact, they improve their own sensorimotor performance through \emph{understanding} the motion goal of their partner (Fig.\ref{f:Ganesh2014}, \cite{Ganesh2014, Takagi2017}). In this paper, we develop an algorithm to replicate this neural mechanism which can be used to improve the sensorimotor performance of a human-robot system.

In order to infer the motion target of a partner, it is necessary to know their control law. However, the robot cannot a-priori know the control used by the human, so instead they must learn it during interaction. This requires the development of an observer through which both partners will understand each other's control in order to predict their motion planning. The paper first describes the design of this observer-predictor pair. It then presents simulation results and an implementation on a robotic interface that exhibit the capabilities of the novel sensory augmentation mechanism.

\section{Problem formulation}
\subsection{System dynamics}
The dynamics of an object manipulated by a robot and a human can be described as
\bea
\label{eq.dynamics}
u + u_h + \epsilon = M \ddot{x} + C \dot{x}
\eea
where $u \! \equiv \! u(t)$ and $u_h \! \equiv \! u_h(t)$ are the robot and human control inputs, respectively, $\epsilon$ is white noise in the robot and human's control inputs, $x \! \equiv \! x(t)$ is the (common) end-effector position in task space, $M \! \equiv \! M(x)$ is the object's mass matrix, and $C\equiv C(x,\dot{x})\dot{x}$ is the Coriolis and centrifugal force term.

We assume that the human and robot generate motions that minimise their error and effort (for the human modelling see \cite{Franklin2008}), corresponding to minimising the respective cost functions
\bea
\label{eq.cost-original}
J_h \!\!\!\!\! &\equiv& \!\!\!\!\! \int_{t_0}^{\infty} \!\!\!\!\! (x-\tau_h)' Q_{h,x}(x-\tau_h) + \dot x' Q_{h,\dot x} \, \dot x + u_h' u_h\, \mbox{d}t \nb \\
J \!\!\!\!\! &\equiv& \!\!\!\!\! \int_{t_0}^{\infty} \!\!\!\!\! (x-\tau)' Q_{x}(x-\tau) + \dot x' Q_{\dot x} \, \dot x + u' u \, \mbox{d}t
\eea
where the subscript $h$ stands for human, $'$ is the transpose operator, $x \!\equiv \!x(t), \, \tau_h \!\equiv \!\tau_h(t), \, \tau \!\equiv \!\tau(t)$ are functions of time, $Q_{h,x}$, $Q_{h,\dot x}$, $Q_{x}$, $Q_{\dot x}$ are positive semi-definite matrices, and $t_0$ is the start time of one trial. $Q_{h,x}$ and $Q_{x}$ are used to express the minimisation of the human and robot's tracking errors, respectively, and $Q_{h,\dot x}$ and $Q_{\dot x}$ the minimisation of their velocity. The weights of the human and robot's control inputs $u$ and $u_h$ are assumed to be 1 for analysis convenience. $\tau_h$ and $\tau$ are the human and robot desired trajectories, respectively, which are unknown to the partner.

To facilitate the analysis, the system dynamics \re{eq.dynamics} and cost functions of the human and robot \re{eq.cost-original} can be written in state-space form as
\bea
\label{eq.system}
\dot \xi \!\!\!\!&=& \!\!\!\!A \, \xi + B (u + u_h + \epsilon) \, , \\
\xi \!\!\!\!\! &\equiv& \!\!\!\!\! \left[ \!\!\! \ba{c} x-\tau \\ \dot{x} \\ x-\tau_h \ea \!\!\! \right]\!\!,\,\,
A \equiv \!\left[\!\!\!
             \begin{array}{cccc}
               0 & 1 & 0 \\
               0 & -M^{-1}C & 0 \\
               0 & 1 & 0 \\
             \end{array}
\!\!\! \right] \!\!,\,\,
B \equiv \left[\!\!\!\!
                               \begin{array}{c}
                                 0 \\
                                 M^{-1} \\
                                 0 \\
                               \end{array}
\!\!\!\! \right]\!\!, \nb \\
\label{eq.cost}
J \!\!\!\!\! &= & \!\!\!\!\! \int_{t_0}^{\infty} \!\!\! \xi' \! (t) \, Q \, \xi(t) + u' \! (t) \, u(t) \, \mbox{d}\tau \nb \\
J_{h} \!\!\!\!\! &= & \!\!\!\!\! \int_{t_0}^{\infty} \!\!\! \xi' \! (t) \, Q_h \, \xi(t) + u_h' \! (t) \, u_h(t) \, \mbox{d}\tau \nb
\eea
where $0$ represents the $n \! \times \! n$ matrix with 0 everywhere, $1$ the $n \! \times \! n$ identity matrix (with 1 as diagonal elements and 0 elsewhere) and
\be
\label{eq.Qh-Q}
Q \equiv \! \left[ \!\!\!\!
      \begin{array}{cccc}
        Q_{x} \!\!& \!\!0 \!\!& \!\!0 \\
        0 \!\!& \!\!Q_{\dot x} \!\!& \!\!0 \\
        0 \!\!& \!\!0\!\! & \!\!0 \\
      \end{array} \!\!
    \right] \!, \,\,\,
Q_h \! \equiv \! \left[\!\!\!\!
      \begin{array}{cccc}
        0 \!\!& \!\!0 \!\!& \!\!0 \\
        0 \!\!& \!\!Q_{h,\dot x} \!\!& \!\!0 \\
        0 \!\!& \!\!0 \!\!&\!\! Q_{h,x} \\
      \end{array} \!\!\!\!
    \right].\nb\\
\ee
In this formulation, both the robot and human use the same state information $\xi$ to minimise their own cost function. Each of them generates motor commands minimising their respective cost function using the LQR algorithm \cite{Kwakernaak1972}:
\bea
\label{e:LQRrobot}
&& u = - L \, \xi \, , \quad L = B' \! P \, , \\
&& A' \! P + P A + Q - P B B' \! P = 0_m \nb \\
\label{e:LQRhuman}
&& u_h = - L_h \, \xi \, , \quad L_h = B' \! P_h \, ,\\
&& A' \! P_h + P_h A + Q_h - P_h B B' \! P_h = 0_m \nb
\eea
where $L$ and $L_h$ are the control gains of the human and robot, respectively, and $P$ and $P_h$ are computed by solving the respective Riccati equation.

\subsection{Sensory augmentation}
Suppose the robot and human's sensing provides them the system's position $x$ and velocity $\dot x$ as well as their own desired trajectory, i.e.
\bea
\label{eq.measurement-human}
&& y_h \equiv \! \left[\!\!\!
          \begin{array}{c}
            \dot x \\
            x-\tau_h \\
          \end{array} \!\!\!
        \right] \!\! + \varepsilon_h
\equiv H_h \, \xi + \varepsilon_h \, , \,\,
H_h \equiv \! \left[ \!\!
            \begin{array}{cccccccc}
              0 & 1 & 0 \\
              0 & 0 & 1 \\
            \end{array} \!\!
          \right] \!, \nb \\
\label{eq.measurement}
&& y \equiv \left[ \!\!\!
          \begin{array}{c}
            x-\tau \\
            \dot x \\
          \end{array} \!\!\!
        \right] \!\! + \varepsilon
= H \xi + \varepsilon \,, \quad H \equiv \left[ \!\!
            \begin{array}{cccccccc}
              1 & 0 & 0 \\
              0 & 1 & 0 \\
            \end{array} \!\!
          \right] .
\eea
where $\varepsilon_h$, $\varepsilon$ represent the respective white measurement noises. How to estimate $\xi$ based on $y_h$ and $y$? In the human-human cooperative tracking task of \cite{Ganesh2014}, the partner's target  information inferred from haptic information was combined with their own visual estimation of the target \cite{Takagi2017}, which resulted in a tracking improvement as shown in Fig.\ref{f:Ganesh2014}. Similarly, could a Kalman filter combining the user and partner's estimated targets be implemented according to their respective noise statistics? In this paper, we apply such a method for the human-robot interaction and design the robot's control input. 

As in the human model of \cite{Takagi2017}, we assume that the two agents estimate each other's desired trajectory and combine it with their own. In particular, the robot can use $\xi \equiv [(x-\tau)'~\dot x'~(x-\hat \tau_h)']'$ to replace the measurement in eq.\re{eq.measurement} where $\hat{\tau}_h$ is the estimate of $\tau_h$. In this way, it is expected that the estimation of the target trajectory will be improved due to the additional sensory signal. We elaborate in the following section how this \emph{sensory augmentation} strategy is realised.

\section{Estimation of human's control}	\label{sec.estimation}
In this section, we develop a method to estimate the human's control input $u_h$ in eq.\re{e:LQRhuman}, which includes two parts unknown to the robot, namely $L_h$ and $\tau_h$. As both of them have to be estimated, we extend the system state from the robot's point of view to
\be
\label{eq.state-extended}
\bar{\xi} \equiv [(x-\tau)'~\dot x'~(x-\tau_h)'~\vec{L}'_{h,x}~\vec{L}'_{h,\dot x}~\vec{L}'_{h,h}]'
\ee
where $\vec{\cdot}$ is the vectorisation operator. The last three components which are from the human's control gain, i.e.
\bea
L_h \equiv [L_{h,x}~L_{h,\dot x}~L_{h,h}]
\eea
correspond to the three variables $x-\tau$, $\dot x$ and $x-\tau_h$, respectively. Then, eqs.(\ref{eq.system},\ref{eq.measurement}) are extended to
\bea
\label{eq.system-extended}
\dot {\bar \xi} \!\!\!&=&\!\!\!\bar{A} \,\bar{\xi} + \bar{B} (u+\epsilon) \,, \\
\bar{A} \!\!\!&\equiv& \!\!\!\!\!\left[\!\!\!
                    \begin{array}{cccccccc}
                      0 \!\!& \!\!1\!\! &\!\! 0\!\! &\!\! 0\!\! & 0 & 0 \\
                      a_{21} \!\!& \!\!a_{22} \!\!& \!\!a_{23} \!\!& \!\!0 \!\!& \!\!0 \!\!& \!\!0 \\
                      0 \!\!& \!\!0 \!\!&\!\! 0\!\! & \!\!0 \!\!& \!\!0 \!\!& \!\!0 \\
                      0 \!\!& \!\!0 \!\!& \!\!0 \!\!& \!\!0 \!\!& \!\!0 \!\!& \!\!0 \\
                      0 \!\!& \!\!0 \!\!& \!\!0\!\! & \!\!0 \!\!& \!\!0 \!\!& \!\!0 \\
                      0 \!\!& \!\!0 \!\!& \!\!0 \!\!& \!\!0 \!\!& \!\!0 \!\!& \!\!0 \\
                    \end{array}\!\!
                  \right]
          \! ,~\bar B \equiv \! \left[\!\!\!
                               \begin{array}{c}
                                 0 \\
                                 M^{-1} \\
                                 0 \\
                                 0 \\
                                 0 \\
                                 0 \\
                               \end{array}\!\!\!\!
                             \right] \nb \\
a_{21} \!\!\!& \equiv & \!\!\!-M^{-1} L_{h,x}\, , \quad a_{22} \equiv -M^{-1}(C+L_{h,\dot x})\,, \nb \\
a_{23} \!\!\! &\equiv & \!\!\! -M^{-1}L_{h,h}\,, \nb \\\nb\\
\label{eq.measurement-1}
y \!\!\! &=& \!\!\! \underline{H} \bar{\xi}+ \varepsilon \, ,\\
\underline{H} \!\!\!&\equiv&\!\!\! \left[\!\!
            \begin{array}{cccccccc}
              1 & 0 & 0 & 0 & 0 & 0 \\
              0 & 1 & 0 & 0 & 0 & 0 \\
              0 & 0 & 1 & 0 & 0 & 0 \\
            \end{array}\!\!
          \right] \! .\nb
\eea
To estimate $L_h$ and $\tau_h$ by using a Kalman filter, we develop the following observer
\bea
\label{eq.observer-extended}
\dot {\hat{\bar{\xi}}}= \hat{\bar{A}} \, \hat{\bar{\xi}} + \bar{B} (u + \epsilon) + K(y - \hat{y}), ~\hat{y} = \underline{H} \hat{\bar{\xi}}
\eea
where $\hat{\bar{A}}$ is the estimate of $\bar{A}$ with $L_h$ replaced by $\widehat{L}_h$, $\hat y$ is the estimate of $y$ and $K$ is the Kalman filter gain. Yielding the estimated extended state $\hat{\bar{\xi}}$, the estimated human's control gain and desired trajectory are obtained.

Note that $\underline{H}$ is a sparse matrix, which indicates that the measurable information of the system is limited. Therefore, it is difficult to simultaneously estimate the human's control gain and desired trajectory. To address this observability issue, we propose to estimate $L_h$ and $\tau_h$ sequentially: the human's control gain $L_h$ is estimated from an initial trajectory $\tau_h(t)$ known to the robot (as can be obtained by asking the human to initially follow a visible target on the robot), after which any human planned trajectory $\tau_h$ can be estimated. The following two subsections describe how $L_h$ and $\tau_h$ can be estimated.

\subsection{Estimation of human's control gain}		\label{sec.gain}
Supposing that the human's initial desired trajectory is known to the robot, the robot's `measurement' yields
\bea
\label{eq.measurement-extended_control}
\bar{y}_1 \!\!\!\!&\equiv& \!\!\!\!\left[
          \begin{array}{c}
            x-\tau \\
            \dot x \\
            x-\tau_h \\
          \end{array}
        \right] + \, \varepsilon_1 \,
= \, \bar H_1 \bar{\xi} \, + \, \varepsilon_1 \, ,\nb\\
\bar{H}_1 \!\!\!\! &\equiv & \!\!\!\! \left[
            \begin{array}{cccccccc}
              1 & 0 & 0 & 0 & 0 & 0 \\
              0 & 1 & 0 & 0 & 0 & 0 \\
              0 & 0 & 1 & 0 & 0 & 0 \\
            \end{array}
          \right]\!.
\eea
Then, $\hat{\bar\xi}$, the robot's estimate of $\bar\xi$,  can be obtained from the observer \re{eq.observer-extended} with the replacements $y \rightarrow \bar{y}_1$ and $\hat{y} \rightarrow \hat{\bar{y}}_1 \equiv \bar {H}_1 \hat{\bar{\xi}}$. The Kalman filter gain is updated iteratively with each time step $k \, \triangle t$ as
\bea
K_1=P_1 \bar H'_1 R_1^{-1}
\eea
where $P_1$ is obtained by solving the Riccati equation
\bea
P_1 \bar A'+ \bar A P_1-P_1\bar H_1'R_1^{-1}\bar H_1P_1+Q_k=0 \, .
\eea
$Q_k$ and $R_1$ are covariance matrices of white noises $\epsilon$ and $\varepsilon_1$, respectively. This minimises the estimation error
\bea
J = E[\|\hat{\bar\xi}-\bar\xi\|^2]
\eea
and the last three components of $\hat{\bar{\xi}}$ form the estimate of $\xi$ used to estimate the human control gain $\widehat L_h$.


\subsection{Estimation of the partner's desired trajectory}\label{sec.target}
With $\widehat{L}_h$ the estimate of the human's control gain $L_h$, it becomes possible to estimate the system state $\xi$ including the human's desired trajectory $\tau_h$, provided that the robot and human reference trajectories are persistently exciting. In particular, the robot's `measurement' including $\widehat{L}_h$ becomes
\bea
\label{eq.measurement-extended}
\bar{y}_2 \!\!\!\!&\equiv& \!\!\!\!\left[
          \begin{array}{c}
            x-\tau \\
            \dot x \\
            \vec{L}_{h,x} \\
            \vec{L}_{h,\dot x} \\
            \vec{L}_{h,h} \\
          \end{array}
        \right] + \, \varepsilon_2 \,
= \, \bar H_2 \bar{\xi} \, + \, \varepsilon_2 \, ,\nb\\
\bar{H}_2 \!\!\!\! &\equiv & \!\!\!\! \left[
            \begin{array}{cccccccc}
              1 & 0 & 0 & 0 & 0 & 0 \\
              0 & 1 & 0 & 0 & 0 & 0 \\
              0 & 0 & 0 & 1 & 0 & 0 \\
              0 & 0 & 0 & 0 & 1 & 0 \\
              0 & 0 & 0 & 0 & 0 & 1 \\
            \end{array}
          \right]\!.
\eea
Similarly as in the previous subsection, $\hat{\bar\xi}$, the robot's estimate of $\bar\xi$, can be obtained from the observer \re{eq.observer-extended} with the replacements $y \rightarrow \bar{y}_2$ and $\hat{y} \rightarrow \hat{\bar{y}}_2 \equiv \bar {H}_2 \hat{\bar{\xi}}$. The Kalman filter gain is updated iteratively with each time step $k \triangle t$ as
\bea
K_2=P_2 \bar H'_2 R_2^{-1}
\eea
where $P_2$ is obtained by solving the Riccati equation
\bea
P_2 \bar A'+ \bar A P_2-P_2\bar H_2'R_2^{-1}\bar H_2 P_2+Q_k=0 \, .
\eea
$R_2$ is covariance matrix of white noise $\varepsilon_2$. Then, the third component of $\hat{\bar{\xi}}$ can be used to obtain the estimate of human's desired trajectory $\tau_h$.

\begin{figure*}[thpb]
      \centering
      \includegraphics[scale=0.252]{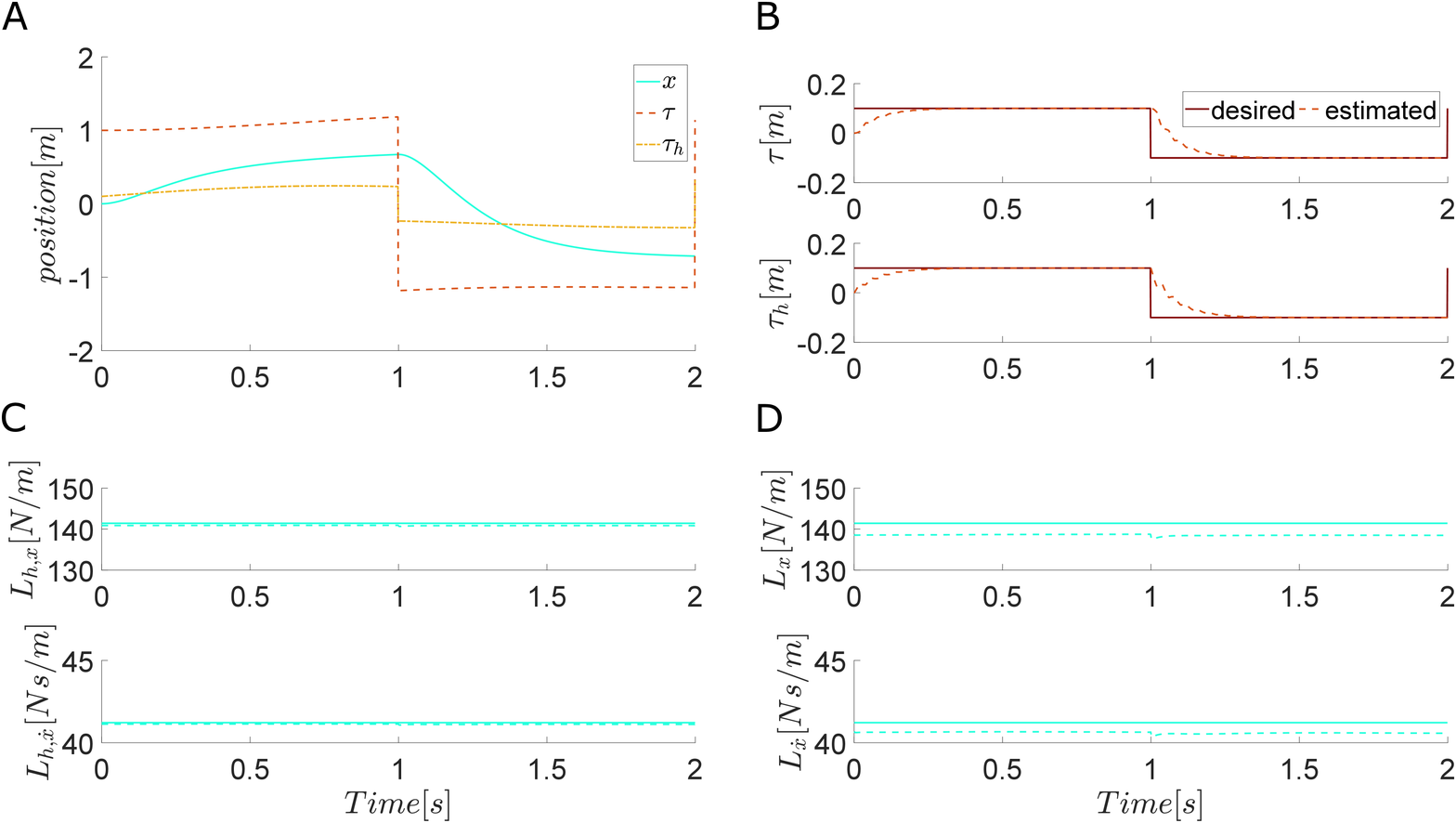}
\caption{Estimation of the partner's control. A: The actual trajectory is in the middle between the human's and robot's desired trajectories, as they have the same weights in their respective cost functions. B: The human and robot are able to estimate each other's planned trajectory. C: Human's control gains $L_{h,x}$, $L_{h,\dot x}$ (solid lines) and their estimates by the robot (dotted lines) almost overlap. D: Robot's control gains $L_x$, $L_{\dot x}$ (solid lines) and their estimates by the human partner (dotted lines) almost overlap.}
      \label{fig.same-weight}
   \end{figure*}

\begin{figure*}[thpb]
      \centering
      \includegraphics[scale=0.252]{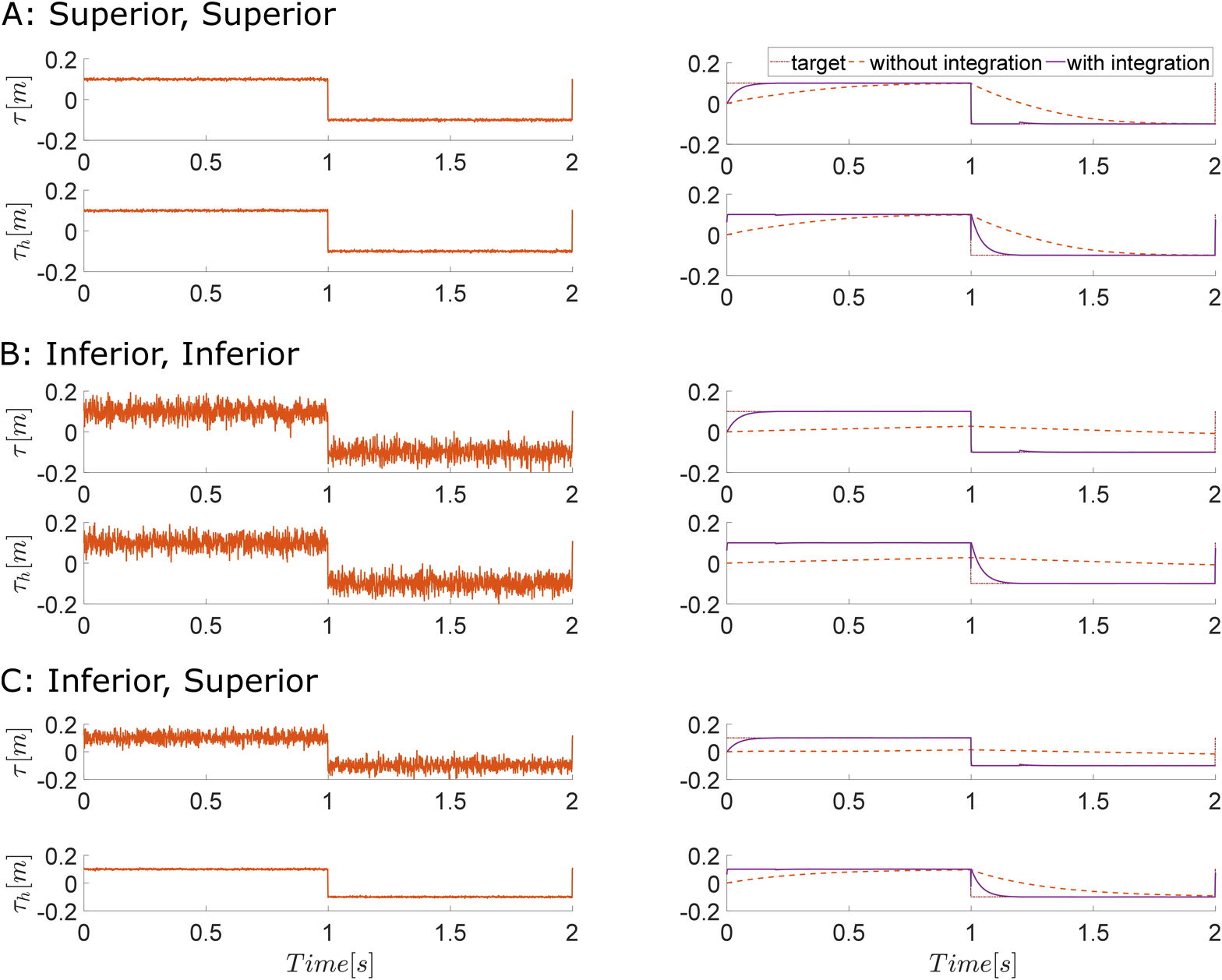}
\caption{Integration of the estimated partner's desired trajectory. The left panels illustrate the measurement of each partner in presence of sensory noise, while the right panels show that target estimation error is reduced for human robot both superior in the tracking task (A), both inferior (B), and with a superior human and an inferior robot (C).}
      \label{fig.integration}
   \end{figure*}

\section{Simulation}
To demonstrate the benefits of the proposed sensory augmentation method, we simulate a scenario where a human arm is rigidly connected to a robot while both track the same target trajectory, i.e. $\tau_h(t) \!=\! \tau(t) \,\, \forall t$. This is simulated by considering the system dynamics Eq. \re{eq.dynamics}, with mass M=6kg and 0 Coriolis and centrifugal component. Motor noise $\epsilon$ is added to the control input (generated using $randn()$ in Matlab). The human and robot use the cost functions of eq.\re{eq.cost-original} with $Q_x=Q_{h,x}=20000$ and $Q_{\dot{x}}=Q_{h,\dot x}=2$.

First, we suppose that the human and robot know each other's initial planned trajectory. The human's desired trajectory is set as a square wave with magnitude of 0.1m and period of 2s, standing for a reaching task including forward and backward movements. The robot's desired trajectory is 1m plus a sweeping signal $[\sin(t) + \sin(2 t) + \sin(3 t)]/10^4$ while the human's desired trajectory is 0.1m plus another sweeping signal $[\cos(t) + \cos(2 t) + \cos(3 t)]/10^4$. Covariance matrices of noises are $Q_k = 10^{-10}1_8$ and $R_1 = 10^{-10}1_4$, respectively. An unscented Kalman filter (UKF) \cite{Wan2000} is implemented to deal with the nonlinearity in the system eq.\re{eq.system-extended} and obtain the estimated control gains $\widehat L$ and $\widehat L_h$, respectively.

Second, the estimated control gains from the first step are used to estimate partner's new desired trajectory. Covariance matrices of noises are $Q_k = 10^{-10}1_8$ and $R_2 = 10^{-10}1_7$, respectively. UKF is implemented again to obtain the estimated desired trajectories $\hat \tau$ and $\hat \tau_h$, respectively.

\subsection{Estimation of the partner's control input}
In this subsection, we first simulate estimation of the partner's control gain. The simulation results are presented in Fig. \ref{fig.same-weight}. Fig.\ref{fig.same-weight}A illustrates the position profile during the reaching task: since human and robot have the same weights in their respective cost functions, the actual trajectory is exactly in the middle between human's and robot's desired trajectories. Fig.\ref{fig.same-weight}C and \ref{fig.same-weight}D show that the human's and the robot's control gains can be reliably estimated by the partner.

With the estimated partner's control gain, we are ready to simulate estimation of the partner's new desired trajectory. In this purpose, we assume that human's and robot's desired trajectories become a square wave with magnitude of 0.1m. Note that they are unknown to the partner. Fig.\ref{fig.same-weight}B illustrates the results of estimation of the partner's desired trajectory. In particular, the upper figure shows that human is able to estimate robot's desired trajectory $\tau$, with a certain error due to continuous change of the movement direction. Correspondingly, the bottom one shows similar performance of estimating human's desired trajectory $\tau_h$ by the robot.


\subsection{Goal integration}
After the partners estimate each other's planned trajectory, they can combine it with their own motion planning. When the two agents track the same target trajectory these two pieces of information can be used to improve the estimation of the `true' target trajectory. To do so the robot uses the `measurement'
\begin{equation}
 y \! \equiv \! [(x-\tau)',\dot{x}',(x-\hat{\tau}_h)']' \!\! + \! \varepsilon.
\end{equation} and the human $y_h \equiv [(x-\hat\tau)',\dot x',(x-\tau_h)']' \! +  \varepsilon_h$. The same target trajectory is set as a square wave with magnitude of 0.1m. Other parameters remain the same as in the previous subsection. 

\begin{figure}[thpb]
      \centering
      \includegraphics[scale=0.252]{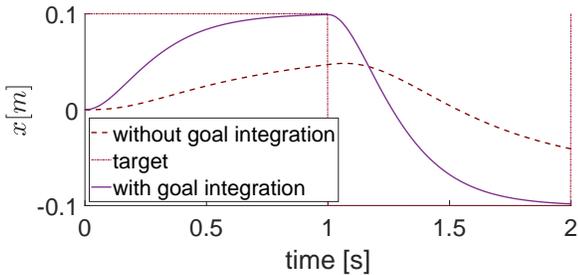}
\caption{Target tracking performance with a superior human and an inferior robot. The performance improvement with integrating the estimated partner's desired trajectory is also found for human robot both superior and both inferior, but is omitted.}
      \label{fig.is_tracking}
   \end{figure}

Fig.\ref{fig.integration} illustrates simulation results with and without integrating the estimated partner's desired trajectory under three conditions:
\begin{itemize}
  \item `Superior' human and superior robot: the covariance matrix of the measurement noise is set as $R_k=10^{-5}$ for both.
  \item `Inferior' human and inferior robot: the covariance matrix of the measurement noise is set as $R_k=10^{-3}$ for both.
  \item Superior human and inferior robot: the covariance matrix of the measurement noise is set as $R_k=10^{-5}$ for the human and $R_k=10^{-3}$ for the robot.
\end{itemize}
It is clear that the estimation performance is improved when the estimated partner's desired trajectory is integrated. Both the human and robot improve performance independent of whether the partner is superior or inferior. These results correspond to the observation of human-human interaction in \cite{Takagi2017}. How about the target tracking performance? Fig.\ref{fig.is_tracking} illustrates that when integrating the estimated partner's desired trajectory, the target tracking performance is improved compared to that without integration. These results show us how the human-robot interaction can be used to improve not only prediction of the target, but also target tracking in collaborative robots.

\section{Experimental Results}
In this section, we implement a similar scenario as in the simulation, where two agents are rigidly connected to a 1 DoF manipulandum. Of the two agents, one comprises the robot, which is estimating its partner, while the other is a `virtual' human agent which acts as a stand in for the human partner. This partner has imposed linear control gains for which the robotic agent is unaware (i.e. Eq. \re{e:LQRhuman}) and thus has to identify. By using a virtual human with known parameters we have a known benchmark and thus can best test the algorithms developed in previous sections.

\begin{figure}[thpb]
      \centering
      \includegraphics[width=1\columnwidth]{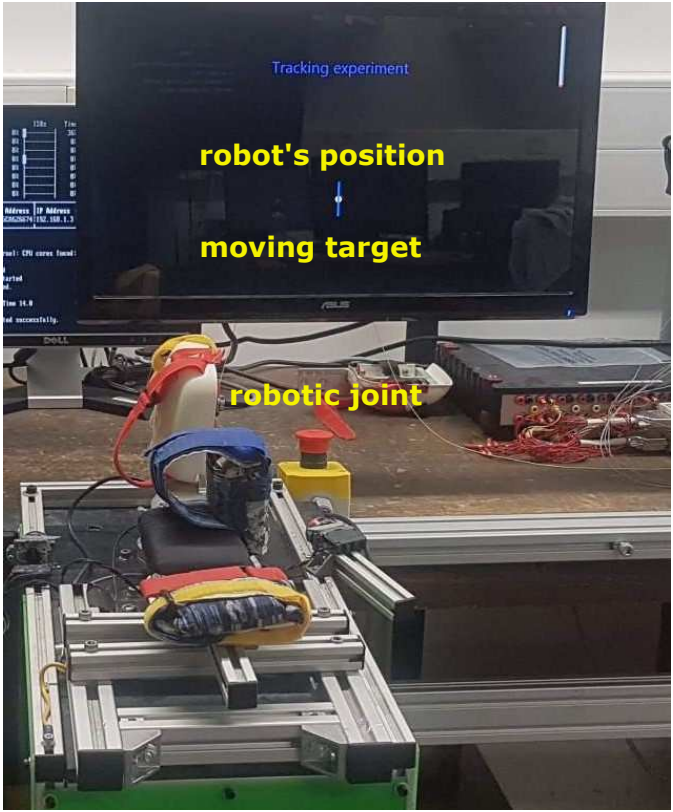}
\caption{Hi5 revolute joint robotic interface. The interface is equipped with a DC motor that allows for application of external torques, e.g., in order to render interaction torques due to physical coupling. The apparatus includes a monitor mounted in front of the manipulandum for providing visual feedback of the robotic joint's actual position and the moving target. This interface is used throughout the implementation results.}
      \label{fig.hi5}
\end{figure}

The experiments are implemented on the Hi5 robotic interface \cite{melendez2011hi5}. Fig.\ref{fig.hi5} depicts this robot which constitutes a 1 DoF revolute joint. After Coulomb and viscous friction compensation, the system can be modeled with the system dynamics Eq. \re{eq.dynamics}, where the inertia is given by $M=0.0035$ $kgm^2$ and there is $0$ Coriolis and centrifugal contribution. The robot motion is controlled by a DC motor and its position is displayed on a monitor. Note that the monitor is not used by the virtual human, who drives the handle with the same motor as the robot although the virtual human control is unknown. The robot agent's component of the control uses the cost functions of eq.\re{eq.cost-original} with $Q_x=1$ and $Q_{\dot{x}}=0$, while the virtual human control is varied.

First, in Section \ref{sec:partner_gains} we suppose that the virtual human and robot know each other's initial planned trajectory. The virtual human's desired trajectory is set as a rounded square wave with magnitude of $8^\circ$ and period of 4s given by
\begin{equation}
    \tau_h(t) = 8\left(sin(0.5\pi t)\right)^{\frac{1}{3}}.
    \label{eqn:human_trajectory}
\end{equation}
This trajectory is chosen as it approximates a point to point reaching task including both forward and backward movements. The robot's desired trajectory is the virtual human's desired trajectory plus a sweeping chirp signal with frequencies ranging from 0Hz to 2Hz to distinguish itself from the virtual human's target. Covariance matrices of noises are $Q_k = 10^{-2}1_6$ and $R_1 = 10^{-7}1_3$, respectively. An unscented Kalman filter (UKF) \cite{Wan2000} is implemented to deal with the nonlinearity in the system eq.\re{eq.system-extended} and to obtain the estimated control gain $\widehat L_h$, which is varied, in different trials, over a range of values.

Second, in Section \ref{sec:partner_trajectory}, one of the estimated control gains from the first experiment is used to estimate the virtual human's now unknown desired trajectory. Covariance matrices of noises are set to $Q_k = 10^{-2}1_6$ and $R_2 = 10^{-7}1_5$, respectively. An UKF is implemented again to obtain the estimated desired trajectory $\hat{\tau}_h$.

Finally in Section \ref{sec:goal_integration}, the estimated virtual human's desired trajectory is used, in a manner consistent with the findings of \cite{Takagi2017}, as additional sensory information about the system. This information is combined with the robot's measurement, using a third UKF, in order to improve the robot's estimate of an uncertain target trajectory.

\subsection{Estimation of the virtual human's control input}
\label{sec:partner_gains}
\begin{figure}[hbtp]
       \centering
       \includegraphics[width=1\columnwidth]{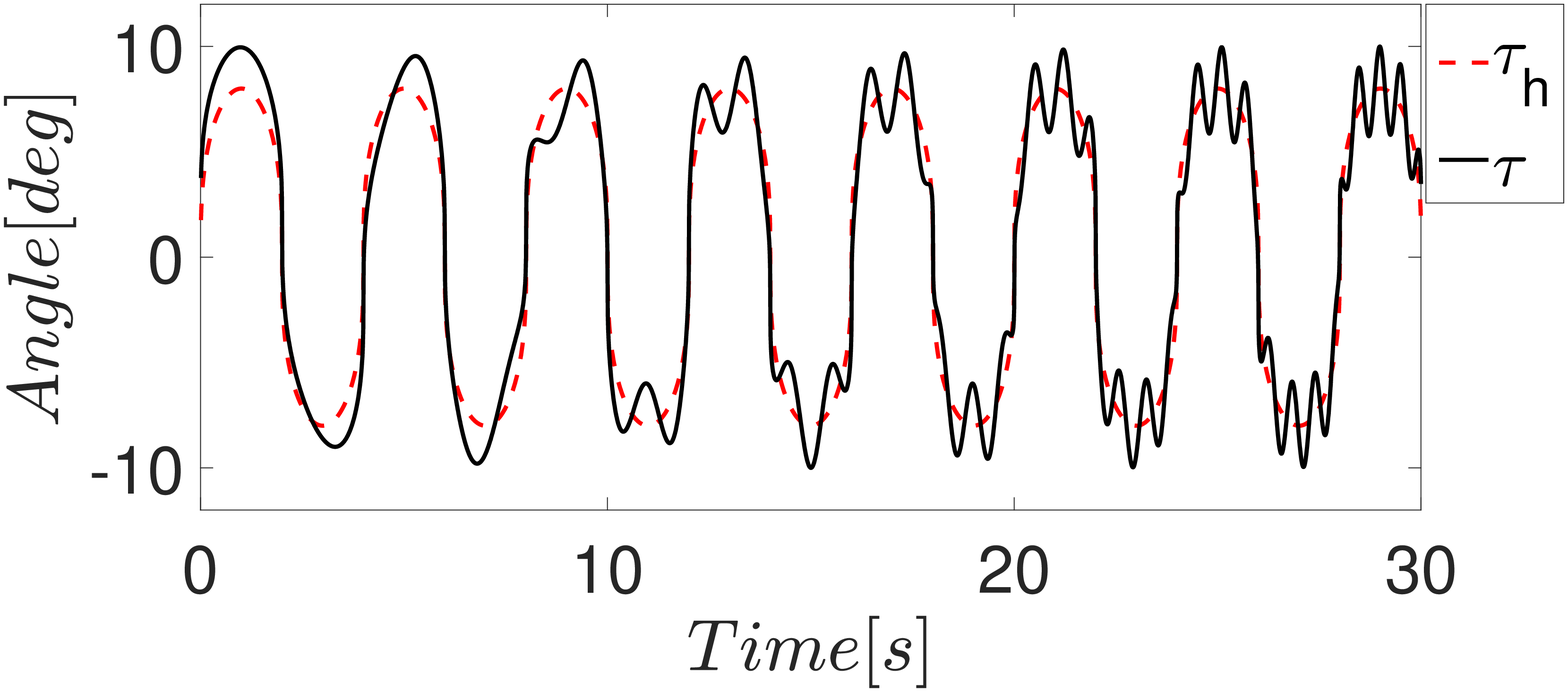}
       \caption{Robot's and virtual human's desired reference trajectories. The red dashed line constitutes the virtual human's reference trajectory $\tau_h$ given by Eq. \re{eqn:human_trajectory} and the black line $\tau$ represents the robot's reference trajectory which has an additional imposed chirp signal.}
       \label{fig:references}
   \end{figure}

\begin{figure}[hbtp]
       \centering
       \includegraphics[width=1.08\columnwidth]{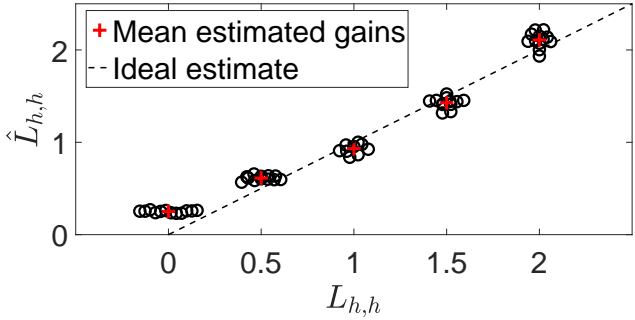}
       \caption{Estimation of the virtual human's varying control gains for 12 trials. The mean estimated values across all trials at the same gain are shown with red crosses, while the individual estimations are represented as black circles.}
       \label{fig:varying_games}
   \end{figure}

In this subsection, we implement estimation of the virtual human's control gain for the known human and robot trajectories (shown in Fig.\ref{fig:references}). With the fixed robot controller gains, we vary the imposed virtual human partner gain from an initial gain of $L_h = [0,0,0]$ to $L_h = [0,0,2]$ in increments of $L_{h,h} = 0.5$. Twelve trials are recorded at each gain value to verify the consistency of the estimations. Fig.\ref{fig:varying_games} shows the resulting partner control gain estimation as a function of the input control gain. Due to the changes in direction of the reference trajectory, these values are reported as the mean value over the final 8 seconds of the interaction. It can be seen that the robot always estimates a value near to the partner's true control gain, however, a small error is present in all cases. This error suggests that even after friction compensation there may be small residual non-linear dynamics. From the figure it can also be observed that the estimation is relatively consistent across trials.  The small observed variation likely results from the probabilistic nature of both the noise and UKFs.

\begin{figure}[hbtp]
       \centering
       \includegraphics[width=1\columnwidth]{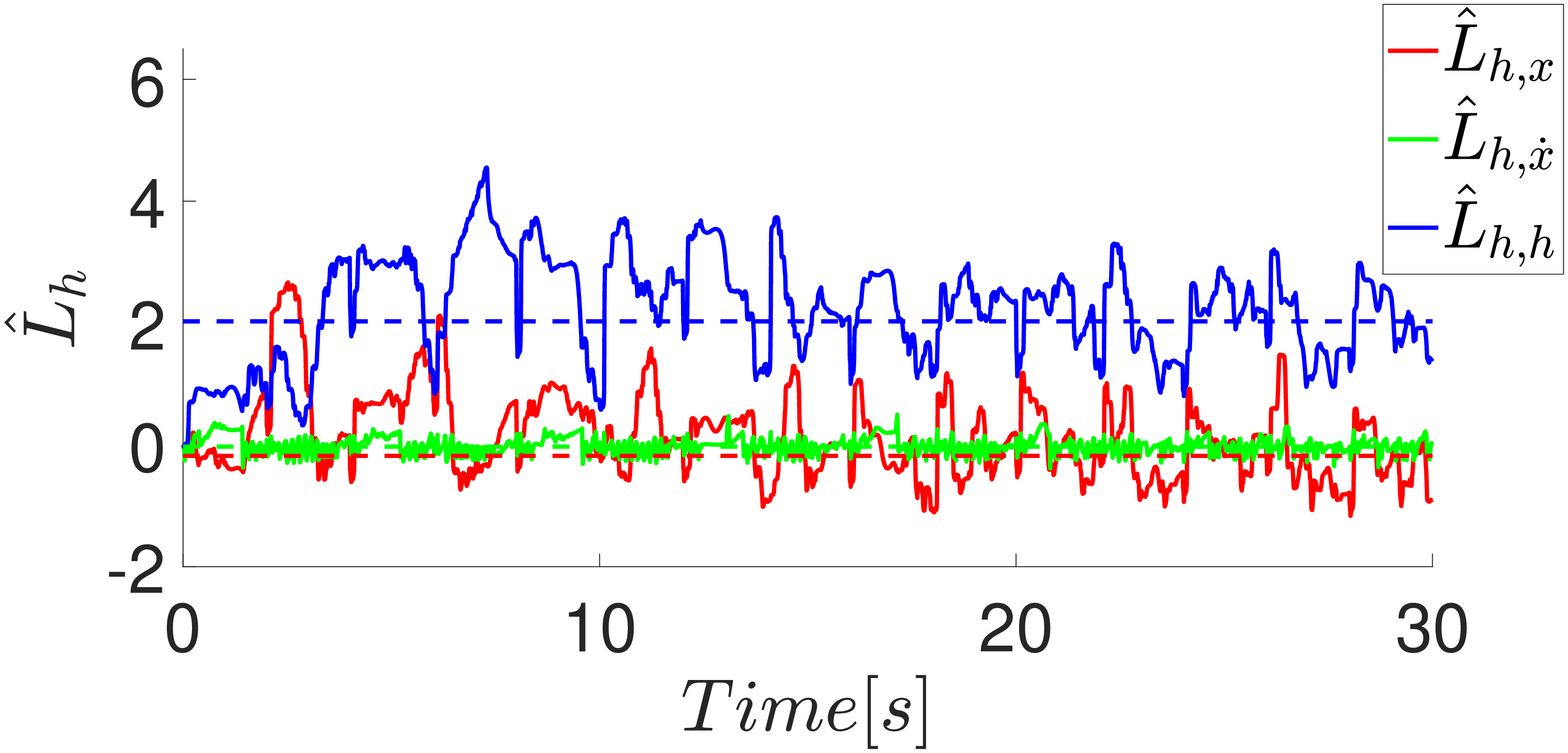}
       \caption{The robot's virtual human gain estimates as a function of time for virtual human gain $L_h = [0,0,2]$. The red curve denotes the estimate of the virtual human partner's robot error gain $\hat{L}_{h,x}$, the green denotes the velocity gain $\hat{L}_{h,\dot{x}}$ and the blue denotes virtual human partner's error gain $\hat{L}_{h,h}$. In each case, the mean value over the last 8 seconds is shown with the coloured dashed line.}
       \label{fig:lhrv}
\end{figure}

To illustrate the convergence behaviour of the implemented UKF, Fig.\ref{fig:lhrv} shows the robot's partner gain estimation as a function of time for the 1st trial with $L_h = [0,0,2]$. It can be seen from this representative example that the estimated gains converge towards an oscillatory behaviour about the true value. This oscillatory behaviour consistently takes place throughout the trajectory and results from the unmodelled change in direction for the reference dynamics.

\subsection{Estimation of the virtual human's reference trajectory}
\label{sec:partner_trajectory}
\begin{figure}[hbtp]
       \centering
       \includegraphics[width=1\columnwidth]{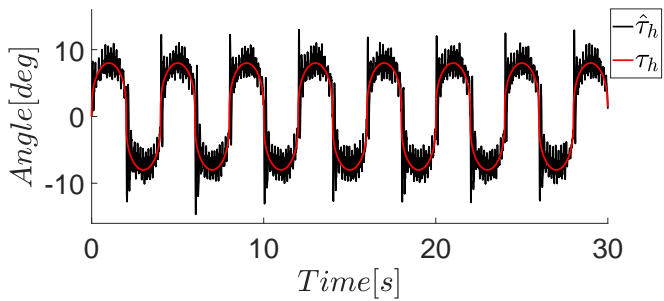}
       \caption{Robot's estimation of the human trajectory. The black line denotes the estimation obtained for a representative trial when $L_{h} = [0,0,2]$, while the red line denotes the true reference trajectory for the virtual human partner.}
       \label{fig:tau_est}
   \end{figure}

With the estimated virtual human's control gain, it is possible to estimate the virtual human's new desired trajectory. In this purpose, we assume that the virtual human and robot's desired trajectories become the same trajectory as given by \eqref{eqn:human_trajectory}. Note that the virtual human trajectory is now unknown to the robotic partner. Fig.\ref{fig:tau_est} illustrates the results of estimation of the partner's desired trajectory. It can be observed that the robot is able to estimate the correct magnitude and shape for the virtual human partner's desired trajectory. However, the estimation possesses a certain amount of error consistent with that observed in the estimated controller gains, likely due to the continuous change of movement direction.


\subsection{Goal integration}
\label{sec:goal_integration}
After the robot estimates the virtual human partner's planned trajectory, it can combine it with its own motion planning. When the two agents track the same target trajectory these two pieces of information can be used to improve the estimation of the `true' target trajectory. To do so the robot uses the `measurement'
\begin{equation}
 y \! \equiv \! [(x-\tau)',\dot{x}',(x-\hat{\tau}_h)']' \!\! + \! \varepsilon.
\end{equation}
The experiment is conducted with the same target trajectory as is given by \eqref{eqn:human_trajectory} and the other parameters remain the same as in the previous subsection. 

\begin{figure}[hbtp]
       \centering
       \includegraphics[width=1.1\columnwidth]{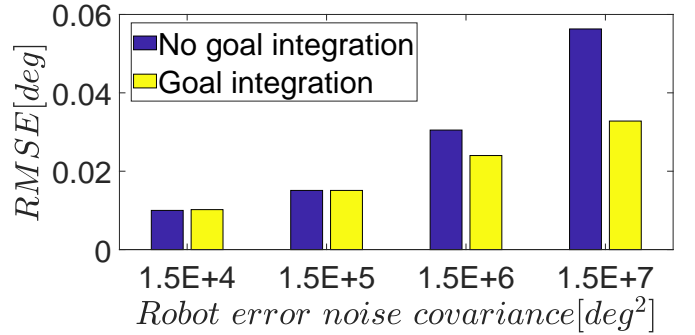}
       \caption{Root mean squared error of the robot error state estimation with and without goal integration. The goal integration provides an improved estimation when the robot's measurement noise level is high.}
       \label{fig:rmse}
   \end{figure}

Fig.\ref{fig:rmse} illustrates the average root mean squared error of the robot's estimated error state with and without integrating the estimated partner's desired trajectory under a range of different injected robot measurement noise levels. When the robot's measurement noise level is relatively high, it is clear that the estimation performance is improved with the estimated virtual human partner's desired trajectory integrated. When the robot's measurement noise level is low, the estimation performance is similar with or without goal integration, as there is not much room to improve the robot's accurate measurement. Together with the simulation results, these experimental results correspond to the observations of human-human interaction in \cite{Takagi2017} and demonstrate that human-robot interaction can be used to improve not only prediction of the target, but also target tracking in collaborative robots.

\section{Discussion}
This paper developed a new algorithm that can explain \emph{haptic communication} between humans. When humans in physical contact have to track the same target, their central nervous system (CNS) estimates each other's planned trajectory, which they integrate with their own visual estimation in order to improve the target trajectory's estimation (Fig.\ref{f:Ganesh2014}, \cite{Takagi2017}). To model this neural mechanism, it is necessary to first identify the partner's control gains and then identify the partner's desired trajectory, which were achieved here through a Kalman filter. After the estimation of the partner's control, a Kalman filter could then combine this information with their own visual observation of the target and plan motion accordingly. Simulation and experimental results showed that this improves the target estimation performance across a range of different interaction noise values.

Importantly, this haptic communication algorithm can be also used to improve the performance of collaborative robots by exploiting the interaction with the user. For instance, when helping a human transporting an object \cite{Agravante2014}, a robot can infer the human's planned movement and so improve its assistance. Similarly, in shared control of semi-autonomous vehicles \cite{Alonso-Mora2014}, the vehicle controller (i.e. the robot) can improve its performance in path tracking using the same strategy. The validity of the developed algorithm in these specific applications and its promising benefits will be explored in future studies. Different from existing works that focused on collaborative control \cite{Jarrasse2014}, this is (in our knowledge) \emph{the first concept and algorithm to use the partner's sensing for improving the robot's sensing and performance}. We note that this algorithm can also be used to optimise the sensing of several interacting robots.

Finally, we have discussed that the observability of the human-robot system dynamics-observation pair is a necessary condition for estimating the partner's control and estimating their motion planning. This condition can be fulfilled if the human and robot exchange rich haptic information. While interaction force was not considered in this paper, it may be used for the simultaneous observation of the partner's control gain and the best motion prediction.


\small
\bibliographystyle{ieeetr}
\bibliography{ref_robot}

\end{document}